\documentclass[10pt,twocolumn,letterpaper]{article}

\usepackage{cvpr}              

\usepackage{graphicx}
\usepackage{amsmath}
\usepackage{amssymb}
\usepackage{booktabs}
\usepackage{multirow}

\usepackage[pagebackref,breaklinks,colorlinks]{hyperref}

\usepackage[capitalize]{cleveref}
\crefname{section}{Sec.}{Secs.}
\Crefname{section}{Section}{Sections}
\Crefname{table}{Table}{Tables}
\crefname{table}{Tab.}{Tabs.}

\def\cvprPaperID{8} 
\def\confName{CVPR}
\def\confYear{2022}

\begin{document}

\title{Combined CNN Transformer Encoder for Enhanced Fine-grained Human Action Recognition}

\author{Mei Chee Leong$^1$, Haosong Zhang$^{1,2}$, Hui Li Tan$^1$, Liyuan Li$^1$, Joo Hwee Lim$^{1,2}$\\
Institute for Infocomm Research (I$^2$R), A*STAR, Singapore$^1$\\
School of Computer Science and Engineering, Nanyang Technological University, Singapore$^2$\\
}


\maketitle

\begin{abstract}
   Fine-grained action recognition is a challenging task in computer vision. As fine-grained datasets have small inter-class variations in spatial and temporal space, fine-grained action recognition model requires good temporal reasoning and discrimination of attribute action semantics. 
   Leveraging on CNN’s ability in capturing high level spatial-temporal feature representations and Transformer’s modeling efficiency in capturing latent semantics and global dependencies, we investigate two frameworks that combine CNN vision backbone and Transformer Encoder to enhance fine-grained action recognition: 
   1) a vision-based encoder to learn latent temporal semantics, and 2) a multi-modal video-text cross encoder to exploit additional text input and learn cross association between visual and text semantics. Our experimental results show that both our Transformer encoder frameworks effectively learn latent temporal semantics and cross-modality association, with improved recognition performance over CNN vision model. We achieve new state-of-the-art performance on the FineGym benchmark dataset for both proposed architectures.

\end{abstract}

\section{Introduction}
\label{sec:intro}

Recognizing actions in video is fundamental and important for video understanding. There are various large-scale action datasets \cite{akarpathy2014sports1m, kay2017kinetics, goyal2017something}, ranging from daily life activities, sports actions, human-object and human-human interactions. These datasets allow increasing research on action recognition, and accelerated the development of deep learning action recognition models. Action classes in coarse-grained datasets, e.g. Kinetics-400
\cite{kay2017kinetics}, can generally be recognized using frame-based approach, as the scene and object information provides strong cue for action prediction. 
On the other hand, fine-grained datasets \cite{goyal2017something, shao2020finegym} have
smaller inter-class variations and requires good temporal
reasoning and domain knowledge.
FineGym \cite{shao2020finegym} is a benchmark dataset of gymnastic videos containing rich details of attribute actions and complex dynamics. 
Each video contains different sequence of attribute actions that represents a distinctive action class, but share similar backgrounds and attributes among video clips.

These fine-grained details pose challenges where accurate classification would depend not only on the visual features but also the latent semantic representations of the attribute actions and their temporal dependencies. Domain expert level text description also plays an important role to discriminate actions that share similar visual representation.

Motivated by such observations, we propose two encoder frameworks to investigate the learning of: 1) discriminative latent temporal semantics in fine-grained actions, and 2) cross-modality association between domain-level text descriptions and attribute visual representations. Convolutional Neural Networks (CNNs) have demonstrated superior performance in vision recognition tasks, with its ability to capture high level visual representations as discriminative features. On the other hand, Transformers (self-attention) \cite{vaswani2017attention}, have been widely used in natural language processing tasks for its key advantage in modeling long-range dependencies. We leverage on the advantages of both models, where we utilize a 3D CNN backbone model to extract high level visual features as sequence of visual tokens, following a Transformer encoder 
to learn latent semantic representations and temporal dependencies.
To exploit additional text modality, we extract vocabulary text from all action class descriptions to form text tokens. Both text and visual tokens are 
passed to a Transformer encoder to learn cross-modality association. The cross encoder framework is trained on action class labels, where the learning of cross visual-text attention is only weakly supervised by the action label.

The main contributions in this paper are summarized as follow: 
(1) we propose a vision encoder framework by combining a 3D CNN vision network and a Transformer encoder for learning latent temporal semantics in fine-grained actions; 
(2) we propose a multi-modal video-text cross encoder framework to exploit additional text input and its association with visual attribute actions by weak supervision on action label;
(3) our experiments show the effectiveness of Transformer encoder in learning latent temporal semantics and cross-modality dependencies for fine-grained actions, producing new state-of-the-art performance on the FineGym benchmark dataset.

\begin{figure*}[t]
\centering 
\includegraphics[width=0.7\textwidth]{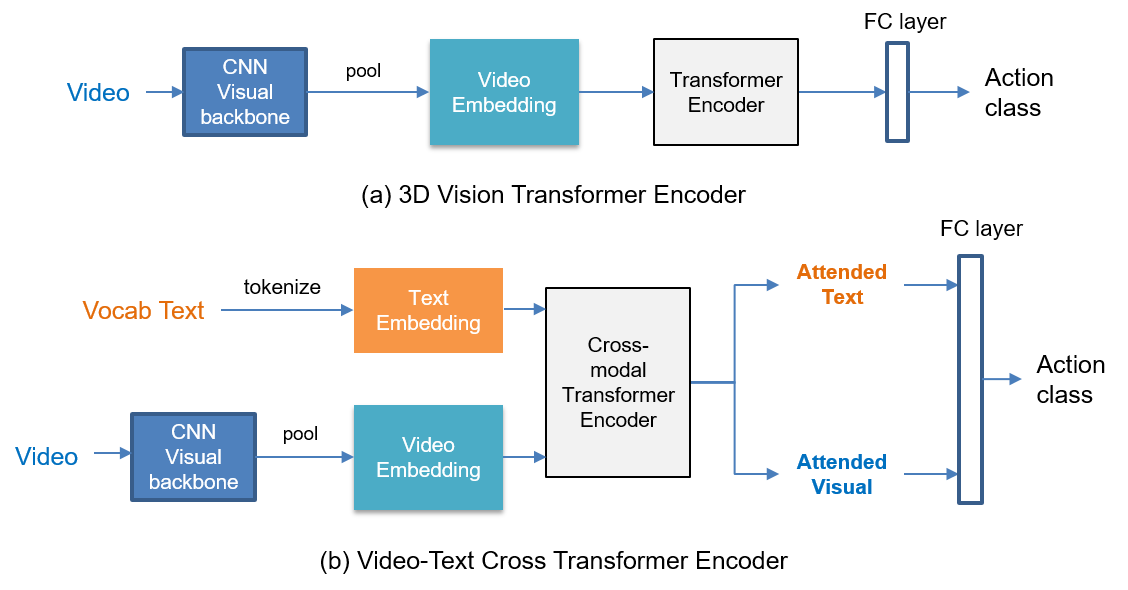} 
\caption{Proposed architectures for enhanced fine-grained action recognition: (a) 3D Vision Transformer Encoder, and (b) Video-Text Cross Transformer Encoder. }
\label{fig:encoder_framework}
\end{figure*}

\section{Related Works}
\label{sec:related_works}

\textbf{Video Action Recognition.} CNN has been the norm for backbone architecture in action recognition, where earlier works start from 2D CNNs \cite{wang2018temporal, zhou2018temporal, lin2019tsm} to extract  frame-based spatial representations following by temporal aggregation. Later works apply 3D CNNs \cite{carreira2017quo, feichtenhofer2019slowfast} directly on consecutive frames to extract spatio-temporal representations.
There are also hybrid frameworks that fuse 2D and 3D CNNs in a network \cite{tran2018closer, leong2020semi, feichtenhofer2020x3d}.
Transformer models have recently been introduced to vision tasks \cite{arnab2021vivit, bertasius2021space}, where each frame is divided into smaller non-overlapped patches to form token representations. These tokens are then embedded before passing to the Transformer blocks. Each Transformer block contains multi-headed self-attention (MHA) \cite{vaswani2017attention}, layer normalization and multi-layer perceptron. Various subsequent works integrate CNN with Transformer \cite{wang2018non, bello2019attention, dai2021coatnet} for complementing CNN and self-attention.
CNN is effective in learning local discriminative visual features by convolution within local neighborhood, while Transformer has strong capacity of learning semantics by self-attention to capture global dependency among tokens. We study the combination of CNN and Transformer to leverage the advantages in learning both discriminative visual features and latent temporal semantics in fine-grained actions. 

\textbf{Video-Text Action Recognition.} TQN \cite{zhang2021temporal} re-formulates fine-grained action recognition task in FineGym as a query-response system, where each response corresponds to a set of attributes. The framework contains a Transformer decoder where the decoder outputs are used for attribute classification and action recognition, providing weak supervision for temporal localization of attributes in untrimmed videos. There are also a number of pre-trained vision-language models \cite{zhu2020actbert, luo2020univl} in natural language processing that can be adopted and fine-tuned for action recognition task.
Compared to these vision-language models, our cross encoder framework does not require ground truth attribute text description for each action class for video-text cross association training and does not require pre-training on large-scale dataset as the video and text are domain-specific.

\section{Proposed Method}
\label{sec:method}

Our proposed CNN Transformer frameworks for 1) 3D vision encoder and 2) video-text cross encoder are presented in 
Fig.~\ref{fig:encoder_framework}. The first architecture employs a 3D CNN vision backbone as feature extractor, followed by a video embedding module. The embedding outputs are passed to a Transformer encoder for learning latent temporal semantics in fine-grained actions. The second architecture extracts visual features using the same 3D vision model and extracts vocabulary set from the action class descriptions as input text tokens. Both visual and text features are fed to respective embedding modules before concatenation and passed to a Transformer-based cross encoder to learn video-text cross-modality attention. The outputs of the cross encoder are subsequently used for action classification.
Both architectures are trained end-to-end with freezed visual backbone, using cross-entropy loss on the action class.

\subsection{3D Vision Transformer Encoder}
\label{ssec:vision_encoder}

An input video is represented as a block of $T \times H \times W \times 3$, with $T$ frames of $H \times W \times 3$ images. For the 3D vision encoder framework, we first extract visual feature maps using a pre-trained 3D vision network.
The dimension of the output features from the vision backbone is denoted as $C^\prime \times T^\prime \times H^\prime \times W^\prime$,  for the channel, temporal and spatial sizes respectively. We apply 2D spatial average pooling on the output of the 3D vision network to obtain feature size $C^\prime \times T^\prime$. 
 
The feature map is passed to the video embedding module for projection to an embedding space of dimension $h=768$ and added with positional embedding information. This position embedding helps to retain temporal order information of the input tokens. The output of the visual embedding of size $T^\prime \times h$ serves as the input for the Transformer encoder. 
We configure the Transformer encoder based on ~\cite{devlin2018bert, luo2020univl}, which consists of encoder hidden layers with MHA, feed-forward networks and layer normalization. The attention mechanism learns contextual relationships among tokens, which is the latent temporal semantics of the fine-grained attribute actions. For our Transformer encoder, we employ three encoder layers to learn high order correspondences between the input tokens.

\subsection{Video-Text Cross Transformer Encoder}
\label{ssec:video_text_encoder}

For our video-text cross encoder framework, we adopt the same network configurations for the visual backbone and video embedding as in the 3D vision encoder framework.  
For the text pathway, we extract the set of attribute texts from FineGym following \cite{zhang2021temporal} as our text input. Let $A = \{a_1, a_2, ..., a_N\}$ denote the set of $N$ attribute text description, where each attribute $a_i$ has varying text length with atomic semantic context. 
We represent each attribute as a single vocabulary with index $i$ as its token representation. 
The vocabulary tokens are then embedded and added with position embedding to obtain text representation with dimension $N \times h$.

To learn cross correspondences of video and text contexts, we pass the concatenated features to a cross embedding layer, followed by a cross-modal Transformer encoder \cite{devlin2018bert, luo2020univl}. In the cross embedding layer, the features embedding is added with positional and token type embedding 
to obtain the cross embedding representation, $(T^\prime + N) \times h$. 
The outputs are then fed to the cross Transformer encoder with two hidden layers to learn the attention of cross-modality.
We split the cross encoder outputs into separate modalities to obtain attended visual and text representations respectively. We perform pooling for individual modality before concatenation for joint prediction. This allows joint learning on the complementary visual and text attended representations.

\section{Experiments}
\label{sec:experiments}

\subsection{Implementation Details}
\label{ssec:implementation}

Our framework is implemented using MMAction2 \cite{zhao2019mmaction} using up to four GPUs. We conduct our experiments on the benchmark FineGym \cite{shao2020finegym} dataset with two settings - 1) Gym99, a more balanced setting with 20k train/8.5k test videos for 99 actions, and 2) Gym288, a long-tailed setting with 23k train/9.6k test videos for 288 actions. The 3D vision encoder is experimented on Gym99, while the video-text cross encoder is evaluated on Gym99 and Gym288.

\subsection{Visual Backbone}
\label{ssec:visual_backbone}

We adopt SlowOnly \cite{feichtenhofer2019slowfast} ResNet-50  as our visual backbone. The model for Gym99 is trained with learning rate 0.01, momentum 0.9, weight decay 1e-4 and gradient clip 40. It is trained with stochastic gradient descent for 120 epochs, with learning rate decay at fixed step 90 and 110. The model for Gym288 has similar training configurations, except for learning rate 0.05 and is trained for 160 epochs, with fixed step at 90 and 140. 
We sample 32 frames with interval 2 as input video, where the image size is $224 \times 224$. After training, we evaluate the model using eight testing clips with single crop. We follow FineGym \cite{shao2020finegym} to evaluate per-video and per-class performances using top-1 accuracy and mean class accuracy.
The results of the SlowOnly visual backbone are shown in Tables \ref{tab:table_1} and \ref{tab:table_2} and serve as our baseline.

\subsection{Experiment on 3D Vision Encoder Network}

To investigate the effectiveness 3D vision encoder in learning latent temporal semantics of attribute actions, we first load and freeze the pre-trained backbone, then train the Transformer encoder end-to-end for 60 epochs using learning rate $3.75\mathrm{e}^{-3}$. 
We evaluate the framework using different number of encoder layers ($B$), and show the results in Table~\ref{tab:table_1}. It is observed that both our 3D vision encoders with single ($B$ = 1) and three hidden layers ($B$ = 3) improve recognition accuracy over the baseline model. Model with three hidden layers performs better than single layer as a higher number of encoder layers helps in learning high order temporal relationships between the latent semantic tokens. By combining SlowOnly with the Transformer encoder, an improvement of 0.55\% in top-1 accuracy is achieved, demonstrating the effectiveness of Transformer encoder in encoding latent temporal semantics for improving fine-grained action recognition. 

\begin{table}[!h]
  \centering
  \caption{Experiment of 3D vision encoder network with different number of encoder layers on Gym99.}
   \begin{tabular}{|l|cc|cc|} \hline
   Network & Top-1 & Mean \\ \hline
   SlowOnly & 93.46 & 89.99     \\
   3D Vision Encoder ($B$ = 1)  &  93.85 & 90.26    \\
   3D Vision Encoder ($B$ = 3)  &  94.01 & 90.54   \\
  \hline
    \end{tabular}%
  \label{tab:table_1}%
\end{table}%

\subsection{Experiment on Video-text Cross Encoder Network}

We investigate the performance of weakly supervised video-text cross association on both Gym99 and Gym288, using attribute text set of size 66 and 98 respectively, that denote the token length for the input text modality.
Our video-text (VT) cross encoder network is trained end-to-end for 30 epochs while freezing the visual backbone.
We adopt AdamW 
as optimizer with learning rate $3\mathrm{e}^{-4}$, weight decay 0.05 and employ Cosine Annealing 
with warm-up strategy as learning rate scheduler. The comparison results with SlowOnly baseline are presented in Table \ref{tab:table_2}. Our VT cross encoder network show improved performance over the baseline visual only model, with significant increments of 1.11$\%$ and 3.28$\%$ in top-1 accuracies for Gym99 and Gym288 respectively. This shows that learning cross attended outputs from separate modalities help the learning of complementary representations from the vision and text semantics, and the weakly supervised video-text association further improve fine-grained action recognition.

\begin{table}[!h]
  \centering
  \caption{Comparison of VT cross encoder performance with SlowOnly baseline on Gym99 and Gym288.}
    \begin{tabular}{|l|cc|cc|} \hline
    \multirow{2}{*}{Network} &  \multicolumn{2}{|c|}{Gym99} & \multicolumn{2}{|c|}{Gym288} \\ \cline{2-5}
     &  Top-1 & Mean & Top-1 & Mean \\ \hline
    SlowOnly &  93.46 & 89.99 & 86.82  & 51.29 \\
    VT Cross Encoder & 94.57 & 91.43 & 90.10 & 62.62 \\
    \hline
    \end{tabular}%
  \label{tab:table_2}%
\end{table}%

\subsection{State-of-the-Art Comparison}

The comparison with existing state-of-the-art (SOTA) performance is listed in Table~\ref{tab:sota}. We compare with the benchmark models used in FineGym, specifically TSN \cite{wang2018temporal}, TRNms \cite{zhou2018temporal} and TSM \cite{lin2019tsm} with 2-stream modalities, I3D \cite{carreira2017quo} and NL I3D \cite{wang2018non} with RGB input. We also compare with a multi-task learning network (MTN) \cite{leong2021joint} that jointly predicts the coarse-to-fine grained action labels and TQN \cite{zhang2021temporal} with video-text query-response system. TSN, TRNms and TSM are pre-trained on ImageNet \cite{deng2009imagenet}, while I3D, NL I3D and TQN are pre-trained on Kinetics-400 \cite{kay2017kinetics}. 
MTN, 3D vision encoder (3D VE) and VT cross encoder (VT CE) are pre-trained on FineGym. 
From the comparison table, both our proposed frameworks outperform existing SOTA in top-1 accuracies for Gym99 and Gym288. 

Our 3D vision encoder network shows significant improvements over existing CNN-based benchmark models with RGB and two stream inputs, demonstrating the advantage of Transformer encoder in learning temporal semantics for fine-grained actions.
Our VT cross encoder network outperforms TQN that trains a query-response system using RGB and text inputs. Our cross-modality network is effective in learning video-text cross encoding, where the learning of visual-text association is weakly supervised by the action class label. VT cross encoder performs better than 3D vision encoder, showing that additional text modality and visual text cross correspondences learn better semantic representations for discriminating attribute actions.
Our framework achieved 94.6$\%$ and 90.1$\%$ in top-1 accuracies for Gym99 and Gym288 respectively, showing increments of 0.8$\%$ and 0.5$\%$ over SOTA.

\begin{table}[!h]
\setlength{\tabcolsep}{1.0pt} 
\small
  \centering
  \caption{Comparison with state-of-the-art performance on FineGym action recognition.}
    \begin{tabular}{|l|l|l|cc|cc|} \hline
    \multirow{2}{*}{Model} & \multirow{2}{*}{Backbone} & \multirow{2}{*}{Modality}  &  \multicolumn{2}{|c|}{Gym99} & \multicolumn{2}{|c|}{Gym288} \\ \cline{4-7}

     &  &  & Top-1 & Mean & Top-1 & Mean \\ \hline
    TSN \cite{wang2018temporal} & BNInception & 2Stream & 86.0 & 76.4 & 79.9 & 37.6      \\
    TRNms \cite{zhou2018temporal} & BNInception & 2Stream & 87.8 & 80.2 & 82.0 & 43.3     \\
    TSM \cite{lin2019tsm} & ResNet-50 & 2Stream & 88.4 & 81.2 & 83.1 & 46.5     \\
    I3D \cite{carreira2017quo} & 3D ResNet-50 & RGB & 75.6 & 64.4 & 66.1 & 28.2 \\
    NL I3D \cite{wang2018non} & 3D ResNet-50 & RGB & 75.3 & 64.3 & 67.0 & 28.0    \\ 
    
    MTN \cite{leong2021joint} & SlowOnly & RGB & 91.8 & 88.5 & - & - \\
    
    TQN \cite{zhang2021temporal} & S3D & RGB+Text & 93.8 & 90.6 & 89.6 & 61.9 \\ \hline
    
    3D VE (Ours) & SlowOnly & RGB & 94.0 & 90.5 & - & - \\
    VT CE (Ours) & SlowOnly & RGB+Text & 94.6 & 91.4 & 90.1 & 62.6 \\

    \hline
    \end{tabular}%
  \label{tab:sota}%
\end{table}%

\section{Conclusion}
\label{sec:conclusion}
In this paper, we 
propose two encoder frameworks that combine CNN and Transformer to investigate learning of latent temporal semantics and cross association of vision text semantics to improve fine-grained human action recognition. Our 3D vision encoder framework allows effective learning of latent temporal semantics from pre-trained CNN visual features. The results of our experiments demonstrate the advantages of exploiting semantic and temporal contexts, which is particularly useful in modeling fine-grained actions where interactions between attributes are very similar among different classes. 
Our video-text cross encoder framework shows effective learning in video-text cross encoding, where the learning of visual-text association is weakly supervised by the action class label. Learning complementary visual text semantics and cross dependencies provide better discrimination for attribute actions that are visually similar. 
Both our proposed encoder frameworks outperform the baseline vision-based model and achieved SOTA performance on the FineGym dataset.
 In future, we plan to investigate integrated CNN and Transformer module designs and different cross encoding configurations. 

\section*{Acknowledgments}
This research is supported by the Agency for Science, Technology and Research (A*STAR) under its AME Programmatic Funding Scheme (Project A18A2b0046).

{\small
\bibliographystyle{ieee_fullname}
\bibliography{egbib}
}

\end{document}